\documentclass{article}

\usepackage{microtype}
\usepackage{graphicx}
\usepackage{subcaption}
\usepackage{booktabs} 

\usepackage{hyperref}

\usepackage[preprint]{icml2026}

\usepackage{amsmath}
\usepackage{amssymb}
\usepackage{mathtools}
\usepackage{amsthm}

\usepackage[capitalize,noabbrev]{cleveref}

\usepackage{graphicx}
\usepackage{wrapfig}
\usepackage{cutwin}
\usepackage{subcaption}         
\usepackage{tabularx} 
\usepackage{array}
\usepackage{booktabs, multirow, caption} 
\usepackage[utf8]{inputenc} 
\usepackage[T1]{fontenc}    
\usepackage{hyperref}       
\usepackage{url}            
\usepackage{amsfonts}       
\usepackage{nicefrac}       
\usepackage{microtype}      
\usepackage[table]{xcolor}  
\usepackage{enumitem}
\usepackage{parskip}
\usepackage{tcolorbox}
\tcbuselibrary{listings}
\usepackage{gensymb}
\definecolor{myblue}{RGB}{216, 233, 246}

\theoremstyle{plain}

\theoremstyle{definition}

\theoremstyle{remark}

\usepackage[textsize=tiny]{todonotes}

\newcommand{\icmlfootnote}{\textsuperscript{*}Work done during the visiting at Australian Institute for Machine Learning, Adelaide University.}

\icmltitlerunning{SpatialNav: Leveraging Spatial Scene Graphs for Zero-Shot Vision-and-Language Navigation}

\begin{document}

\twocolumn[
  \icmltitle{SpatialNav: Leveraging Spatial Scene Graphs for Zero-Shot Vision-and-Language Navigation}
  


  \icmlsetsymbol{footnote}{*}

  \begin{icmlauthorlist}
    \icmlauthor{Jiwen Zhang}{fdu,aiml,footnote}
    \icmlauthor{Zejun Li}{fdu}
    \icmlauthor{Siyuan Wang}{usc}
    \icmlauthor{Xiangyu Shi}{aiml}
    \icmlauthor{Zhongyu Wei}{fdu,sh}
    \icmlauthor{Qi Wu}{aiml}
  \end{icmlauthorlist}

  \icmlaffiliation{fdu}{Fudan University, Shanghai, China}
  \icmlaffiliation{usc}{University of Southern California, Los Angeles, USA}
  \icmlaffiliation{aiml}{Australian Institute for Machine Learning, Adelaide University, Australia}
  \icmlaffiliation{sh}{Shanghai Innovation Institute, Shanghai, China}
  \icmlcorrespondingauthor{Zhongyu Wei}{zywei@fudan.edu.cn}


  \vskip 0.3in
]



\printAffiliationsAndNotice{\icmlfootnote}  

\begin{abstract}
  Although learning-based vision-and-language navigation (VLN) agents can learn spatial knowledge implicitly from large-scale training data, zero-shot VLN agents lack this process, relying primarily on local observations for navigation, which leads to inefficient exploration and a significant performance gap. To deal with the problem, we consider a zero-shot VLN setting that agents are allowed to fully explore the environment before task execution. Then, we construct the \textbf{Spatial Scene Graph (SSG)} to explicitly capture global spatial structure and semantics in the explored environment. Based on the SSG, we introduce \textbf{SpatialNav}, a zero-shot VLN agent that integrates an agent-centric spatial map, a compass-aligned visual representation, and a remote object localization strategy for efficient navigation. Comprehensive experiments in both discrete and continuous environments demonstrate that SpatialNav significantly outperforms existing zero-shot agents and clearly narrows the gap with state-of-the-art learning-based methods. Such results highlight the importance of global spatial representations for generalizable navigation.
\end{abstract}

\section{Introduction}
\label{sec:intro}

Vision-and-Language Navigation (VLN)~\cite{anderson2018vision,krantz2020beyond} is a fundamental problem in the embodied AI community, which requires an agent to follow natural language instructions to navigate in complex real-world environments. Early works in VLN mainly focused on supervised learning by designing dedicated model architectures~\cite{hong2021vln,qiao2022hop,chen2022think} or introducing better training strategies~\cite{wang2019reinforced,zhang2021curriculum}. More recently, the success of Large Language Models (LLMs) has inspired the adoption of multimodal large language models (MLLMs) to construct zero-shot VLN agents, such as NavGPT~\cite{zhou2024navgpt} and Open-Nav~\cite{qiao2025open}. Although these zero-shot agents demonstrate good generalization capabilities without task-specific training, there is still a significant performance gap compared to learning-based methods. 

\begin{figure}[t]
\setlength{\abovecaptionskip}{-1pt}
\centering
    \includegraphics[width=\linewidth]{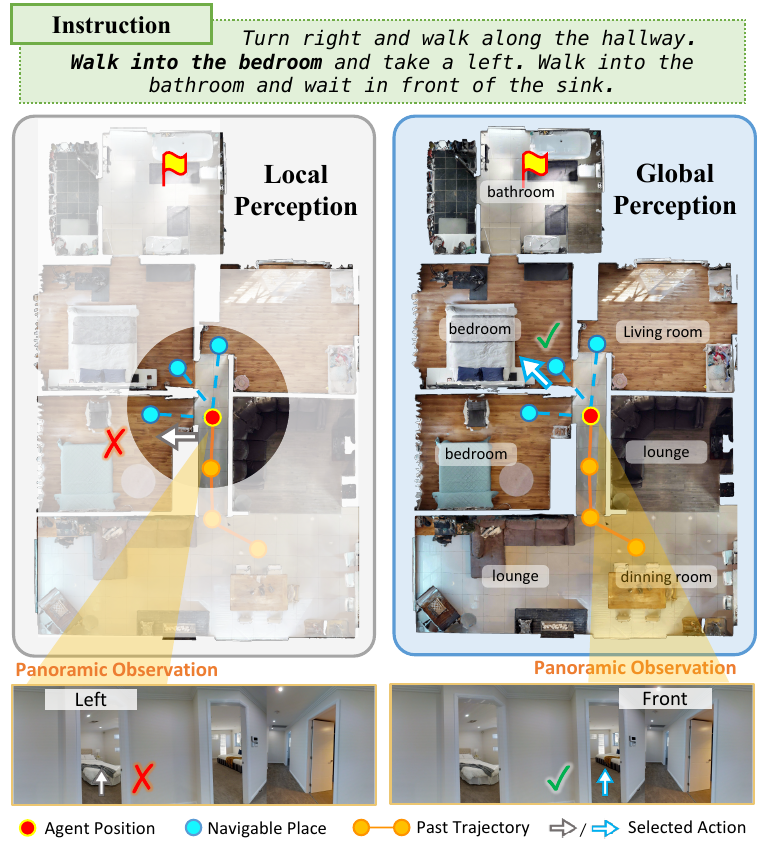}
    \caption{\textbf{Illustration of local v.s. global perception during navigation.} 
    When the instruction mentions a \textit{\underline{bedroom}}, agent maybe confused by local perception if multiple bedrooms are present and plausible. In contrast, global spatial information enables the agent to disambiguate these options and make more accurate actions.
    }
    \label{fig:intro}
\end{figure}

The key factor contributing to this disparity lies in how zero-shot agents perceive the environment. Unlike learning-based VLN agents which can implicitly acquire the regularities about room layouts and functional co-occurrences (e.g. kitchens often connect to dining areas, and closets are usually adjacent to bedrooms) through large-scale pre-training~\cite{chen2021history,wang2023scaling,zhou2024navgpt2}, zero-shot agents typically rely on constrained local observations for decision making without access to such spatial priors~\cite{zhou2024navgpt,chen2024mapgpt,shi2025smartway,he2025strider}. Therefore, navigation decisions are often short-sighted, leading to inefficient exploration. Moreover, as shown in Figure~\ref{fig:intro}, under such a local perception mechanism, navigation becomes particularly challenging when multiple instruction-consistent actions co-exist, as the agent lacks sufficient global information to disambiguate among them. However, how to enable zero-shot agents to access and use global spatial information for effective navigation remains a fundamental challenge.

To resolve the challenge, we propose a new zero-shot VLN setting that allows agents to fully explore the environment before task execution. This setting is grounded in practical considerations, since in real-world household applications such as vacuum cleaners, robots are deployed in bounded indoor environments and rarely switch scenes. 
While it departs from the original definition of VLN that restricts agents to purely online local perception, this setting reflects realistic deployment scenarios and opens up new opportunities for utilizing global spatial information.
Under this setting, we propose to equip zero-shot VLN agents with global perception ability by constructing a \textbf{Spatial Scene Graph~(SSG)} that explicitly captures global spatial layouts and semantics in the environment. 
Since environment reconstruction through pre-exploration using off-the-shelf SLAM systems~\cite{liu2025slam3r} is feasible, we design an automatic pipeline to organize the reconstructed point cloud into a hierarchical graph structure. This pipeline involves segmenting floors and rooms, annotating room categories and detecting objects with labels. Our spatial scene graph serves as a compact knowledge base of the environment, enabling the agent to conduct long-horizon reasoning beyond local observations. 

To effectively apply the spatial scene graph for navigation, we design \textbf{SpatialNav}, a zero-shot VLN agent with three novel components. (1) Since a spatial scene graph typically contains far more information than required for a specific navigation task, we propose to construct \textbf{\textit{an agent-centric spatial map}} by querying the SSG with the current location of the agent and retrieving the task-relevant spatial layout within a bounded region. (2) We propose to organize the panoramic observations into \textbf{\textit{a compass-like visual representation}}, ensuring consistent orientation with the spatial map. (3) To support future-aware decision making, we further introduce \textbf{\textit{remote object localization}}, which queries the SSG to retrieve the objects around candidate navigable locations, enabling the agent to reason about future observations beyond its current field of view. Extensive experiments in both discrete and continuous environments validate the superiority of SpatialNav, which achieves a success rate of 57.7\%, 49.6\%, 64.0\%, 32.4\% on val-unseen splits of R2R, REVERIE, R2R-CE, RxR-CE, respectively, surpassing all previous zero-shot agents and even several supervised-learning methods. 

In conclusion, our contributions are:
\begin{itemize}[leftmargin=*,partopsep=0pt,topsep=0pt]
\setlength{\itemsep}{0pt}
\setlength{\parsep}{0pt}
\setlength{\parskip}{0pt}
    \item We introduce a novel zero-shot VLN setting that allows agents to pre-explore the environment, and construct the \textbf{Spatial Scene Graph (SSG)} to efficiently encode global spatial structure and semantic information from pre-exploration.
    \item We design \textbf{SpatialNav}, a zero-shot VLN agent that leverages the spatial scene graph through agent-centric spatial map, compass-like visual representation, and remote object localization.
    \item We empirically prove that global spatial information can be effectively utilized by different agents and across different environments. Our results indicate that zero-shot VLN agents equipped with spatial scene graphs can achieve comparable performance with SOTA learning-based agents. 
\end{itemize}

\begin{figure*}[t]
\setlength{\abovecaptionskip}{-2pt}
\centering
    \includegraphics[width=\textwidth]{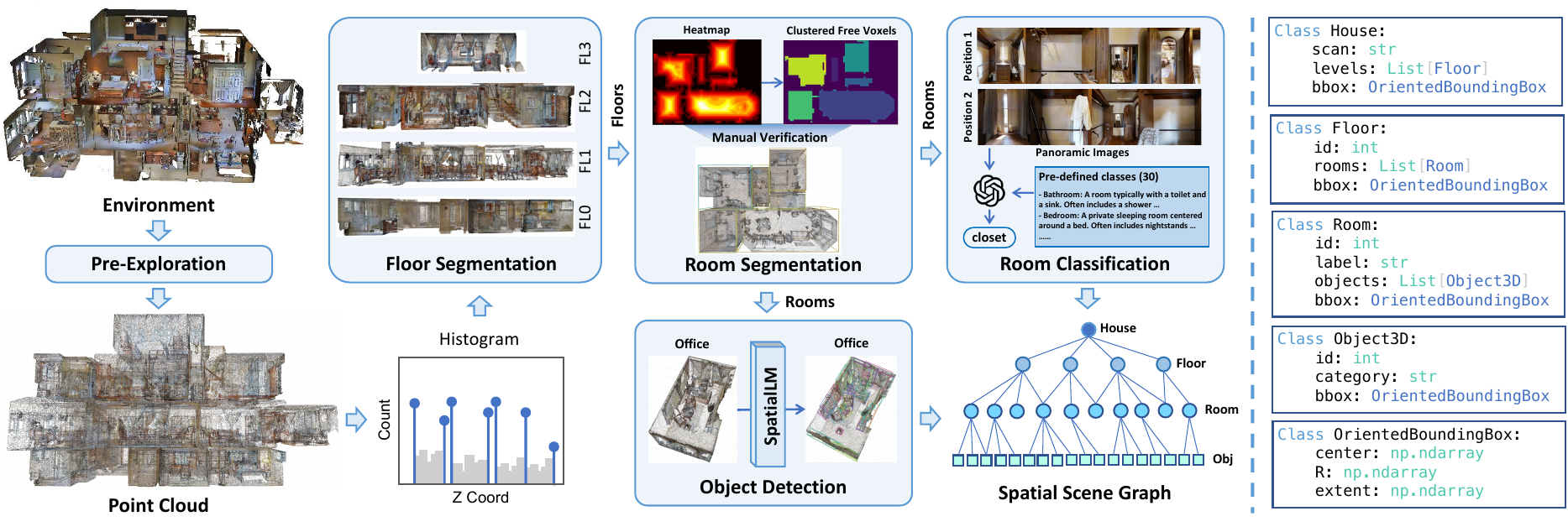}
    \caption{\textbf{Overview of the spatial scene graph construction.} Given the point cloud input, we employ a four-stage annotation pipeline, including floor segmentation via height-based clustering, room segmentation within each floor using geometric heuristics, room classification based on visual observations and object detection from room-level point clouds. The definitions of node entities in the spatial scene graph are illustrated on the right.}
    \label{fig:framework}
\end{figure*}

\section{Related Works}
\label{sec:related_works}

\paragraph{Vision-Language Navigation (VLN).} VLN~\cite{anderson2018vision, krantz2020beyond, ku2020room} is one of fundamental challenges in embodied AI. 
Early studies predominantly focused on discrete environments~\cite{qi2020reverie,jain2019stay,zhu2021soon}, where the action space is abstracted as a predefined navigation graph. For better adaptation to realistic open-world scenarios, \cite{krantz2020beyond,hong2022bridging} has extended VLN to continuous environments~\cite{savva2019habitat} that allow low-level control. Under both settings, supervised training~\cite{tan2019learning,hao2020towards,zhu2020vision,zhang2021curriculum,chen2021history,wang2023scaling,zhang2025embodied,du2024delan,zhang2025activevln} on annotated datasets is the mainstream paradigm for advancing the performance, especially when combined with MLLMs~\cite{zhang2024navid,wei2025streamvln,zheng2025efficient}. 
However, these learning-based VLN agents have significant reliance on large-scale, domain-specific training, limiting their generalization to unseen environments.

\paragraph{Zero-Shot VLN Agents.} Recent works have investigated zero-shot VLN agents~\cite{long2024discuss,qiao2025open,chen2025constraint} that leverage (M)LLMs for navigation. These methods typically rely on complex prompting strategies, such as action-aware reasoning~\cite{zhou2024navgpt,chen2023A2}, deliberative planning~\cite{long2025instructnav}, progress estimation~\cite{he2025strider} and mistake reflection~\cite{shi2025smartway}. 
Despite these advances, these agents still lag behind the learning-based agents by a clear margin. Therefore, we propose Spatial-Nav, a strong zero-shot VLN agent that achieves competitive performance with SOTA supervised learning-based agents. 

\paragraph{Environment Exploration in VLN.} Previous learning-based agents have incorporated environment exploration as a useful tool to refine the navigation policy, either by self-supervised learning on successful exploration trajectories~\cite{wang2019reinforced} or by training a topological map planner~\cite{chen2021topological}. Recently, VL-KnG~\cite{mdfaa2025vl} employs pre-explored videos to construct object-centric knowledge graphs for goal identification, proving that exploration can compensate for the absence of learned spatial priors. However, such pre-exploration settings have not been systematically validated for zero-shot VLN agents. Although VLN-Zero~\cite{bhatt2025vln} takes a step in this direction by building a symbolic scene graph after exploration, its exploration remains limited as it focuses only on symbolic constraints, overlooking the spatial layouts and semantics that are critical for VLN.
To mitigate this gap, we propose to fully explore the environment and to build spatial scene graphs that integrate global layouts with room and object semantics, enabling efficient and generalizable navigation.

\begin{figure*}[t]
\setlength{\abovecaptionskip}{1pt}
\centering
    \includegraphics[width=\textwidth]{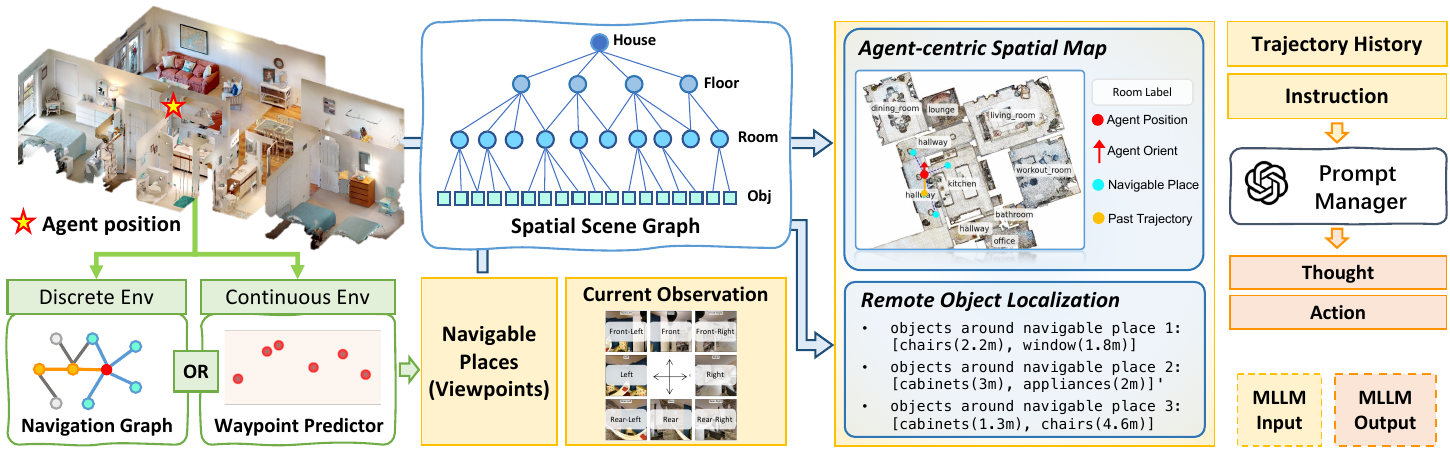}
    \caption{\textbf{The framework of SpatialNav agent.} 
    Based on its current position, SpatialNav queries the SSG to construct an agent-centric spatial map
    and to retrieve the object semantic descriptions around navigable places. 
    Panoramic visual observations are organized into a single compass-like image. These information, combined with trajectory history and language instructions, constitutes the context of SpatialNav for predicting the next action.}
    \label{fig:spatialnav}
\end{figure*}

\section{Methods}
\label{sec:methods}

\paragraph{Problem setup.} Following real-world applications such as robotic vacuum cleaners, we adopt a VLN setting where an agent is allowed to pre-explore the environment before executing the tasks. Since dense or sparse point clouds can be efficiently obtained using existing SLAM systems~\cite{liu2025slam3r,murai2025mast3r, maggio2025vggt} by taking RGB videos, we make the practical assumption that the 3D point cloud of the environment is available after exploration. 
Given the reconstructed 3D structure provided by Matterport3D dataset~\cite{chang2017matterport3d}, we focus on two key objectives:
Firstly, to enable spatial-enhanced navigation, we endow the raw 3D point cloud with structured spatial annotations (e.g., floor–room–object hierarchy) and semantic labels (e.g., room types and object categories), constructing a spatial scene graph that can be reused across different episodes.
Secondly, we propose a SpatialNav agent to efficiently utilize the resulting spatial scene graph. 

\subsection{Spatial Scene Graph (SSG)}
\label{subsec:spatial_scene_graph}

We construct the spatial scene graph by supplementing structural and semantic annotations on the raw point cloud, as demonstrated in Figure~\ref{fig:framework}. The process is divided into four stages: (1) \textbf{Floor Segmentation}: Following~\cite{werby2024hierarchical}, we segment the floors by calculating a height histogram of all points, applying DBSCAN and selecting the highest-ranking peaks as the floors. (2) \textbf{Room Segmentation}: Then, we perform room-level segmentation within each floor. We adopt the geometric heuristic-based method proposed by~\cite{bobkov2017room}, which formulates room segmentation as a partitioning problem of enclosed regions. However, we find this approach relies heavily on the presence of strong geometric boundaries, such as walls, to delineate room regions. As a result, its performance degrades when multiple functional areas share a continuous open space. To resolve such problems, we refine the room segmentation results by manually verifying the regions with areas larger than 20 square meters. (3) \textbf{Room Classification}: To assign semantic labels to each segmented room, we collect the images or video frames captured within each room during the pre-exploration phase and prompt GPT-5~\cite{openai2025gpt5systemcard} to classify the room according to a predefined room category list. (4) \textbf{Object Detection}: To detect objects within each room, we fine-tune SpatialLM~\cite{mao2025spatiallm}, a model that consumes the segmented room point clouds to predict the object bounding boxes with labels, on the training scans of Matterport3D~\cite{chang2017matterport3d}. 

Finally, we organize floors, rooms, and objects into a hierarchical spatial scene graph, where nodes correspond to entities (listed on the right side of Figure~\ref{fig:framework}) and edges represent the containment relations between different levels of the hierarchy. 

\subsection{SpatialNav Agent}
\label{subsec:spatial_nav}

Typically, VLN agents make navigation decisions based on local visual observations, the instruction, and navigation history, predicting the next action at each step~\cite{zhou2024navgpt,shi2025smartway,he2025strider}. However, such a local perception field constrains the agent's performance, as it lacks awareness of the spatial layout and semantic cues of the future. Equipped with a spatial scene graph, we resolve the limitations by designing an MLLM-based VLN agent, namely \textbf{SpatialNav}. SpatialNav distinguishes itself with previous (M)LLM-based agents from three distinct perspectives: (1) an agent-centric spatial map that captures the layout of surrounding rooms to support coarse-grained spatial reasoning, (2) a compass-like visual observation schema that aligns egocentric images with the orientation of the spatial map, and (3) a remote object localization strategy, which retrieves object-level semantic information around navigable places from the spatial scene graph to guide the fine-grained grounding. The framework of SpatialNav is summarized in Figure~\ref{fig:spatialnav}.

\paragraph{Agent-centric Spatial Map.} Previous LLM-based VLN agents are mostly restricted to a local perception range within about 3 meters away from their current position~\cite{zhou2024navgpt, chen2024mapgpt, qiao2025open}. Different from them, we enlarge the perception range by querying the spatial scene graph to generate an agent-centric spatial map. Specifically, given the position of the agent, we determine the floor level based on its Z-axis coordinate, and then identify the room it resides in according to its X-Y coordinates within that floor. After localizing the agent, we define an agent-centric spatial receptive field with a radius of approximately 7 meters within the same floor. The agent position and nearby rooms are then projected onto a top-down spatial map, with the agent heading always aligned to the upward direction. By offering a compact and structured representation of the scene, the spatial map allows the agent to reason about its immediate spatial context without being overwhelmed by irrelevant information.

\paragraph{Compass-like Visual Observation.} Prior zero-shot VLN agents, such as SpatialGPT~\cite{jiang2025spatialgpt} and Smartway~\cite{shi2025smartway}, usually process panoramic observations by capturing multiple images and feeding them sequentially into the MLLM for action prediction. However, this strategy results in high input token overhead. In contrast, we discretize the panorama into eight directional views (from 0\degree ~to 360\degree, turning right 45\degree ~each) with a field of view set as 90\degree ~to avoid adjusting elevations. Instead of feeding the visual observations of eight views sequentially, we organize them into a single image with a compass-style representation. As shown in the lower part of Figure~\ref{fig:spatialnav}, this compass image is arranged as a 3×3 grid, where the eight views are placed along the perimeter of the grid in clockwise order. The center of this image is occupied by a compass that explicitly encodes the agent relative orientation with respect to these views. This schema ensures consistency between the egocentric observations with the spatial map and reduces the cost of visual inputs. 

\paragraph{Remote Object Localization.} To further help with the future planning, we propose to provide the agent with the awareness of ``which objects will be observed if I took that direction''. Specifically, for each candidate navigable place obtained either from the navigation graph in discrete environments or via a waypoint predictor in continuous environments, we query the spatial scene graph to retrieve the object semantics within a local neighborhood of that place. The retrieved information includes object categories and their distances from that navigable place is then compressed as a concise textual description, added into the context of SpatialNav.

By incorporating these components together, SpatialNav enables long-horizon, goal-aware decision making that goes beyond local perception.

\begin{table*}[t]
\centering
\small
\captionsetup{skip=2pt}
\setlength{\tabcolsep}{2pt}
\setlength{\belowcaptionskip}{2pt}
\caption{\textbf{Comparison with SOTAs in discrete environments on the R2R and REVERIE Val-Unseen splits.} The \textbf{best} and the \underline{second best} results are denoted by \textbf{bold} and \underline{underline}. ${}^{\dagger}$ means the spatial annotations are ground truth.}
\label{tab:discrete_main}
\begin{tabular}{l|c|ccccc|ccc}
\toprule
\multirow{2}{*}{\textbf{Settings}} & \multirow{2}{*}{\textbf{Methods}} & \multicolumn{5}{c}{\textbf{R2R}} & \multicolumn{3}{|c}{\textbf{REVERIE}} \\
\cmidrule{3-10}
&  & \textbf{TL} & \textbf{NE(↓)} & \textbf{OSR(↑)} & \textbf{SR(↑)} & \textbf{SPL(↑)} & \textbf{OSR(↑)} & \textbf{SR(↑)} & \textbf{SPL(↑)} \\
\midrule
\multirow{6}{*}{\begin{tabular}[c]{@{}l@{}} Supervised\\ Learning \end{tabular}}
& NavCoT~\cite{lin2024navcot} & 9.95 & 6.36 & 48 & 40 & 37 & 14.2 & 9.2 & 7.2 \\
& PREVALENT~\cite{hao2020towards} & 10.19 & 4.71 & - & 58 & 53 & -- & -- & -- \\
& VLN-BERT~\cite{hong2021vln} & 12.01 & 3.93 & 69 & 63 & 57 & 27.7 & 25.5 & 21.1 \\
& HAMT~\cite{chen2021history} & 11.46 & \underline{2.29} & 73 & 66 & \underline{61} & 36.8 & 33.0 & 30.2 \\
& DUET~\cite{chen2022think} & 13.94 & 3.31 & \underline{81} & \underline{72} & 60 & \underline{51.1} & \underline{47.0} & \underline{33.7} \\
& DUET+ScaleVLN~\cite{wang2023scaling} & 14.09 & \textbf{2.09} & \textbf{88} & \textbf{81} & \textbf{70} & \textbf{63.9} & \textbf{57.0} & \textbf{41.8} \\
\midrule
\multirow{6}{*}{\begin{tabular}[c]{@{}l@{}} Zero-Shot \end{tabular}}
& NavGPT~\cite{zhou2024navgpt} & 11.45 & 6.46 & 42 & 34 & 29 & 28.3 & 19.2 & 14.6 \\
& MapGPT~\cite{chen2024mapgpt} & -- & 5.63 & 57.6 & 43.7 & 34.8 & 36.8 & 31.6 & 20.3 \\
& MC-GPT~\cite{zhan2024mc} & -- & 5.42 & 68.8 & 32.1 & -- & 30.3 & 19.4 & 9.7 \\
& SpatialGPT~\cite{jiang2025spatialgpt} & -- & 5.56 & \textbf{70.8} & 48.4 & 36.1 & -- & -- & -- \\ 
& \cellcolor{myblue} SpatialNav (ours) & \cellcolor{myblue}13.8  & \cellcolor{myblue}\underline{4.54} &  \cellcolor{myblue}\underline{68.2} &  \cellcolor{myblue}\underline{57.7} & \cellcolor{myblue}\underline{47.8} & \cellcolor{myblue}\textbf{58.1} & \cellcolor{myblue}\underline{49.6} & \cellcolor{myblue}\underline{34.6} \\
& \cellcolor{myblue} SpatialNav${}^{\dagger}$ (ours) & \cellcolor{myblue}13.8 &  \cellcolor{myblue}\textbf{4.40} & \cellcolor{myblue}\textbf{70.8} & \cellcolor{myblue}\textbf{59.3} & \cellcolor{myblue}\textbf{48.0} & \cellcolor{myblue}\underline{57.8}	&  \cellcolor{myblue}\textbf{50.4}	& \cellcolor{myblue}\textbf{33.7}  \\
\bottomrule
\end{tabular}
\end{table*}
\begin{table*}[t]
\centering
\small
\captionsetup{skip=2pt}
\setlength{\tabcolsep}{2pt}
\setlength{\belowcaptionskip}{3pt}
\caption{\textbf{Comparison in continuous environments on the R2R-CE and RxR-CE Val-Unseen splits.} The \textbf{best} and the \underline{second best} results are denoted by \textbf{bold} and \underline{underline}. ${}^{\dagger}$ means the spatial annotations are ground truth.}
\label{tab:continuous_main}
\resizebox{\linewidth}{!}{
\begin{tabular}{l|c|ccccc|cccc}
\toprule
\multirow{2}{*}{\textbf{Settings}} & \multirow{2}{*}{\textbf{Methods}} & \multicolumn{5}{c}{\textbf{R2R-CE}} & \multicolumn{4}{|c}{\textbf{RxR-CE}} \\
\cmidrule{3-11}
&  & \textbf{NE(↓)} & \textbf{OSR(↑)} & \textbf{SR(↑)} & \textbf{SPL(↑)} & \textbf{nDTW(↑)} & \textbf{NE(↓)} & \textbf{SR(↑)} & \textbf{SPL(↑)} & \textbf{nDTW(↑)} \\
\midrule
\multirow{7}{*}{\begin{tabular}[c]{@{}l@{}} Supervised\\ Learning \end{tabular}}
& MapNav~\cite{zhang-etal-2025-mapnav} & 4.93 & 53.0 & 39.7 & 37.2 & -- & 7.62 & 32.6 & 27.7 & 43.5 \\
& VLN-BERT~\cite{hong2021vln} & 5.74 & 53.0 & 44.0 & 39.0 & -- & 8.98 & 27.1 & 22.7 & 46.7 \\
& GridMM~\cite{wang2023gridmm} & 5.11 & 61.0 & 49.0 & 41.0 & -- & -- & -- & -- & -- \\
& ETPNav~\cite{an2024etpnav} & 4.71 & 65.0 & 57.0 & 49.0 & -- & 5.64 & 54.8 & 44.9 & 61.9 \\
& HNR~\cite{wang2024lookahead} & \underline{4.42} & 67.0 & 61.0 & 51.0 & -- & 5.51 & 56.4 & 46.7 & 63.6 \\
& NavFoM~\cite{zhang2025embodied} & 4.61 & \underline{72.1} & \underline{61.7} & \underline{55.3} & -- & \underline{4.74} & \underline{64.4} & \textbf{56.2} & \underline{65.8} \\
& Efficient-VLN~\cite{zheng2025efficient} & \textbf{4.18} & \textbf{73.7} & \textbf{64.2} & \textbf{55.9} & -- & \textbf{3.88} & \textbf{67.0} & \underline{54.3} & \textbf{68.4} \\
\midrule
\multirow{7}{*}{\begin{tabular}[c]{@{}l@{}} Zero-Shot \end{tabular}}
& OpenNav~\cite{qiao2025open} & 6.70 & 23.0 & 19.0 & 16.1 & 45.8 & -- & -- & -- & -- \\
& CA-Nav~\cite{chen2025constraint} & 7.58 & 48.0 & 25.3 & 10.8 & -- & 10.37 & 19.0 & 6.0 & -- \\
& Smartway~\cite{shi2025smartway} & 7.01 & 51.0 & 29.0 & 22.5 & -- & -- & -- & -- & -- \\
& STRIDER~\cite{he2025strider} & 6.91 & 39.0 & 35.0 & 30.3 & 51.8 & 11.19 & 21.2 & 9.6 & 30.1 \\
& VLN-Zero~\cite{bhatt2025vln} & 5.97 & 51.6 & 42.4 & 26.3 & -- & 9.13 & 30.8 & 19.0 & -- \\
& \cellcolor{myblue} SpatialNav (ours)
& \cellcolor{myblue}\underline{5.15} & \cellcolor{myblue}\underline{66.0} & \cellcolor{myblue}\underline{64.0} & \cellcolor{myblue}\underline{51.1} & \cellcolor{myblue}\underline{65.4}
&\cellcolor{myblue}\underline{7.64} & \cellcolor{myblue}\underline{32.4} &  \cellcolor{myblue}\underline{24.6} & \cellcolor{myblue}\underline{55.0} \\
& \cellcolor{myblue} SpatialNav${}^{\dagger}$ (ours) 
& \cellcolor{myblue}\textbf{4.21} & \cellcolor{myblue}\textbf{73.0} & \cellcolor{myblue}\textbf{68.0} &\cellcolor{myblue}\textbf{53.4} & \cellcolor{myblue}\textbf{69.3} 
& \cellcolor{myblue}\textbf{7.34} & \cellcolor{myblue}\textbf{39.0} & \cellcolor{myblue}\textbf{28.4} & \cellcolor{myblue}\textbf{56.0} \\
\bottomrule
\end{tabular}}
\end{table*}

\section{Experiments}
\label{sec:experiment}

\subsection{Experimental Setup}

\paragraph{Datasets and Simulators} We evaluate our approach in discrete and continuous environments. For discrete environment, we use the R2R~\cite{anderson2018vision} and REVERIE~\cite{qi2020reverie} datasets with Matterport3D simulator~\cite{chang2017matterport3d}. The validation unseen split for R2R and REVERIE contains 783 and 1328 trajectories spanning 11 scans. For continuous environment, we utilize the R2R-CE~\cite{krantz2020beyond} and RxR-CE~\cite{ku2020room} datasets with Habitat v0.3.2 simulator~\cite{savva2019habitat}. Following previous zero-shot VLN agents in continuous environments~\cite{qiao2025open, shi2025smartway, he2025strider}, we randomly sampled a subset of 100 and 200 trajectories from the validation unseen split of R2R-CE and RxR-CE, respectively.  

\paragraph{Evaluation Metrics} In our experiments, we report the widely used standard VLN evaluation metrics~\cite{hong2021vln,wang2023scaling,zhang2025embodied,qiao2025open}, including trajectory length (TL), navigation error (NE), success rate (SR), oracle success rate (OSR), success weighted by path length (SPL) and normalized Dynamic Time Warping (nDTW). An episode is considered successful if the agent stops within 3 meters of the goal in both discrete and continuous environments. 

\paragraph{Baselines} 
In discrete environment, we compare against six supervised learning-based agents and four zero-shot agents, where ScaleVLN~\cite{wang2023scaling} and SpatialGPT~\cite{jiang2025spatialgpt} achieve the state-of-the-art performance.
In continuous environment, we compare against seven supervised learning-based agents and five zero-shot agents, where Efficient-VLN~\cite{zheng2025efficient} and VLN-Zero~\cite{bhatt2025vln} achieve the state-of-the-art performance within each group.

\begin{table*}[t]
\centering
\small
\captionsetup{skip=2pt}
\setlength{\tabcolsep}{2pt}
\setlength{\abovecaptionskip}{2pt}
\setlength{\belowcaptionskip}{3pt}
\caption{\textbf{Comparison with baselines enhanced with spatial map.} ``SMap'' is short for agent-centric spatial map. To ensure environmental diversity, we use the 56 scans on the validation splits of R2R and REVERIE, based on which we sample 267 and 230 trajectories, respectively. For R2R-CE, we use the randomly sampled 100 trajectories.}
\label{tab:ablation_spatial}
\begin{subtable}[t]{0.55\textwidth}
\centering
\resizebox{\linewidth}{!}{
\begin{tabular}{l|ccc|ccc}
\toprule
\multirow{2}{*}{\textbf{Method}} & \multicolumn{3}{c}{\textbf{R2R-Val-Sampled}} & \multicolumn{3}{|c}{\textbf{REVERIE-Val-Sampled}} \\
\cmidrule{2-7}
& \textbf{OSR(↑)} & \textbf{SR(↑)} & \textbf{SPL(↑)} & \textbf{OSR(↑)} & \textbf{SR(↑)} & \textbf{SPL(↑)} \\
\midrule
SMap Only & 61.8 & 40.8 & 31.7 & 59.5 & 33.2 & 17.7 \\
\midrule
NavGPT & 53.2 & 43.5 & 34.7 & 37.1 & 31.5 & 23.4 \\
NavGPT + SMap & 63.7 & 52.1 & 42.7 & 57.8 & 47.4 & 34.1 \\
\midrule
SpatialNav & \textbf{68.9} & \textbf{60.3} & \textbf{50.1} & \textbf{56.5} & \textbf{50.9} & \textbf{34.3} \\
\bottomrule
\end{tabular}}
\caption{Results on discrete environments.}
\label{tab:sub_spatial_discrete}
\end{subtable}
\hfill
\begin{subtable}[t]{0.4\textwidth}
\centering
\resizebox{0.91\linewidth}{!}{
\begin{tabular}{l|ccc}
\toprule
\multirow{2}{*}{\textbf{Method}} & \multicolumn{3}{c}{\textbf{R2R-CE Val-Unseen}} \\
\cmidrule{2-4}
 & \textbf{OSR(↑)} & \textbf{SR(↑)} & \textbf{SPL(↑)} \\
\midrule
SmartWay & 60.0 & 51.0 & 41.0 \\
SmartWay + SMap & \textbf{68.0} & 62.0 & 50.3 \\
\midrule
SpatialNav & 66.0 & \textbf{64.0} & \textbf{51.1} \\
\bottomrule
\end{tabular}}
\caption{Results on continuous environments.}
\label{tab:sub_spatial_continuous}
\end{subtable}
\end{table*}


\begin{figure*}[t]
\setlength{\abovecaptionskip}{-1pt}
\centering
    \includegraphics[width=\textwidth]{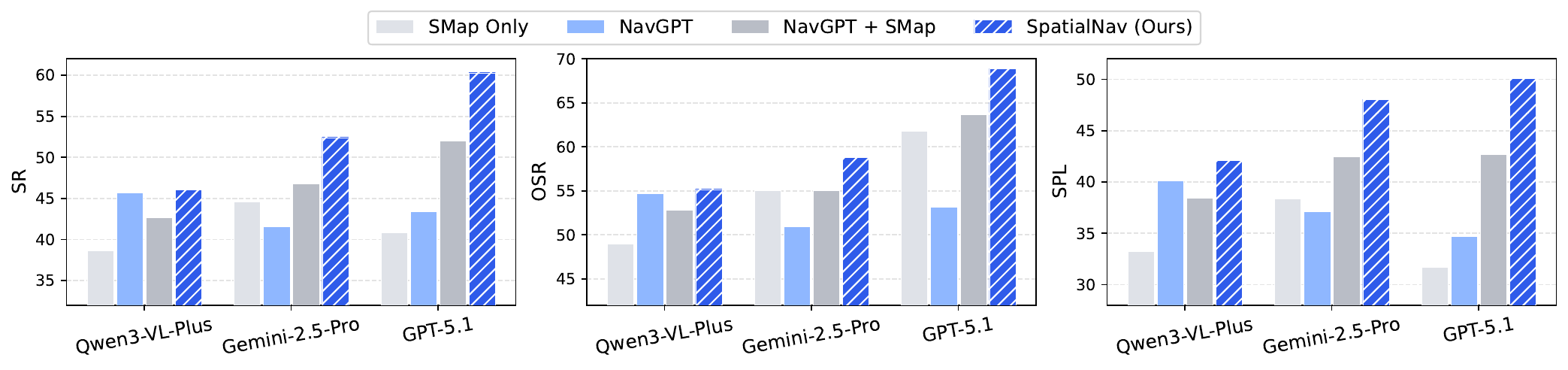}
    \caption{\textbf{Comparison between different MLLM backbones.} We report the SR, OSR and SPL. }
    \label{fig:llm_ablation}
\end{figure*}

\paragraph{Implementation Details} As we have stated in Section~\ref{subsec:spatial_nav}, we divide the panoramic observations into eight directions, representing the front, front-right, right, rear-right, rear, rear-left, left and front-left. The 256 $\times$ 256 images taken at each view are organized into a single 1024 $\times$ 1024 compass-like image. For the spatial map, we set its width and height as 1024, with gird size set as 0.015m, leading to a perception radius of 7.68m. For remote object localization, we retrieve the objects within the same room of the navigable places and group them together based on the predicted object categories. Note that in continuous environment, we utilize the waypoint predictor proposed by~\cite{shi2025smartway} to obtain the navigable places. we use GPT-5.1~\cite{openai2025gpt5systemcard} as the MLLM backbone.

\subsection{Main Results}

We compare SpatialNav with other state-of-the-art VLN agents under two different settings: supervised learning and zero-shot. The results in discrete and continuous environments are summarized in Table~\ref{tab:discrete_main} and Table~\ref{tab:continuous_main}, respectively. Note that we also report the performance of a variant of our method, namely SpatialNav${}^{\dagger}$, where the spatial annotations about room segmentations, room types and object detections are replaced with ground-truth annotations provided by the Matterport3D dataset~\cite{chang2017matterport3d}, serving as an upper-bound reference. 

In discrete environments, SpatialNav consistently outperforms existing zero-shot baselines on both R2R and REVERIE val-unseen splits. Compared with the strongest zero-shot baseline, SpatialGPT, SpatialNav achieves an absolute gain of +9.3\% in SR and +11.7\% in SPL on R2R val-unseen, demonstrating stronger navigation effectiveness and path efficiency. Although SpatialNav navigates in a zero-shot manner, its performance is comparable with several supervised learning-based agents, such as PREVALENT and VLN-BERT on the R2R dataset and DUET on the REVERIE dataset.

In continuous environments which pose additional challenges due to low-level control, SpatialNav significantly outperforms prior zero-shot methods on both R2R-CE and RxR-CE. In particular, SpatialNav significantly outperforms VLN-Zero, an approach closely related to us since it also leverages a form of global memory constructed through pre-exploration. While VLN-Zero only maintains a symbolic scene graph that lacks explicit semantic structure, SpatialNav builds a spatial scene graph that jointly models global layout and detailed semantics. As a result, SpatialNav yields absolute gains of +21.6\% SR and +24.8\% SPL on R2R-CE. Performance gains are also observed on the more challenging RxR-CE dataset, indicating that spatial knowledge can is beneficial for VLN. 

Moreover, SpatialNav${}^{\dagger}$ further improves the performance across four datasets in both discrete and continuous environments, demonstrating that the quality of spatial annotations plays a critical role in navigation performance. This observation encourages future work on more accurate and scalable automatic spatial annotation methods for VLN.

\subsection{Further Analysis}

In this section, we analyze the soundness of SpatialNav by addressing the following two questions:

\paragraph{Q1: Does explicit spatial knowledge serve as an effective signal for VLN?} 
Firstly, in discrete environments where navigable viewpoints are stored as navigation graphs and are easily accessible, we design a spatial map only (SMap Only) baseline that navigates solely on the agent-centric spatial map, the navigation instruction, and the set of navigable locations, without using any visual or textual observations of the panorama. 
As shown in Table~\ref{tab:sub_spatial_discrete}, this baseline achieves good performance, validating that spatial maps alone contain sufficient spatial information to support navigation decisions. Then, we evaluate whether the spatial map provides consistent benefits when incorporated into existing VLN agents. Specifically, we augment NavGPT~\cite{zhou2024navgpt} in discrete environments and SmartWay~\cite{shi2025smartway} in continuous environments with the spatial map as additional input, while keeping their original mechanisms unchanged. Results in Table~\ref{tab:ablation_spatial} demonstrate that explicit spatial information constitutes an effective and generalizable signal for VLN across different agents and environment settings.

Furthermore, we evaluate two alternative multimodal large language models (MLLMs) to explore whether the effectiveness depends on the choice of MLLM backbones. We select another strong closed-source model, Gemini-2.5-Pro~\cite{comanici2025gemini}, and a representative open-source model Qwen3-VL-Plus~\cite{bai2025qwen3vltechnicalreport}. Results are summarized in Figure~\ref{fig:llm_ablation}. Both GPT-5.1 and Gemini-2.5-Pro exhibit clear gains when augmented with spatial maps, whereas Qwen3-VL-Plus-based NavGPT slightly degrades when augmented with the spatial map. We observe that the output pattern of Qwen3-VL-Plus is highly repetitive, suggesting that it may be constrained by prior fine-tuning on VLN-style data. For instance, thinking patterns such as “\textit{Thought: I have reached/moved/entered/exited} ...” account for approximately 30\% of the outputs, while patterns like “\textit{Thought: The instruction requires} ...” appear in about 20\% of the cases. Despite these differences, SpatialNav consistently achieves the best performance across all evaluated backbones, indicating that our method is largely backbone-agnostic. 

\paragraph{Q2: How do different components of SpatialNav affect the navigation?} We conduct a progressive analysis by incrementally enriching a minimal baseline along the dimensions of density and quality of spatial information, together with the modality of panoramic observation. We evaluate on 50 trajectories on 5 validation scans with the average recall and precision of room segmentation exceeding 50\% to ensure a reasonable quality. We implement the minimal baseline as a text-based agent that utilizes the descriptions of each viewpoint from~\cite{zhou2024navgpt} with no spatial information. 

As shown in Table~\ref{tab:ablation_perception}, applying the spatial map improves the performance. 
However, adding remote object semantics under the text-only setting leads to performance degradation, particularly with ground-truth annotations. This is caused by semantic ambiguity between objects described in textual panoramic observations and those retrieved from future locations, which hinders accurate termination decisions and results in longer trajectories and higher navigation error. Such ambiguity is more severe for ground truth as it contains many small objects that cannot be predicted by SpatialLM. Importantly, this issue is alleviated when replacing the textual panoramic observations with our compass-style visual observations. We find that visual grounding helps to align current perceptions with future spatial cues. This indicates richer spatial semantics are most effective when combined with visual observations. Besides, we can also observe that ground-truth annotations consistently outperform model-predicted ones, while predicted annotations still provide clear benefits. Overall, these results validate the design of SpatialNav.


\subsection{Ablation Studies}
Moreover, we conduct experiments on the sampled subset of R2R validation splits spanning 56 scans to validate the hyper-parameter settings of SpatialNav.

\begin{table}[t]
\centering
\small
\captionsetup{skip=2pt}
\setlength{\tabcolsep}{1.5pt}
\caption{\textbf{Impact of the density and quality of spatial information on sampled subset of R2R validation scans.} ``G'' and ``P'' indicate that the spatial annotations are ground truth and model-predicted, respectively.}
\label{tab:ablation_perception}
\resizebox{\linewidth}{!}{
\begin{tabular}{c|cc|cc|ccccc}
\toprule
\multirow{2}{*}{\textbf{\begin{tabular}[c]{@{}c@{}}Pano\\ Obs\end{tabular}}} & \multicolumn{2}{|c}{\textbf{\begin{tabular}[c]{@{}c@{}}Spatial\\ Map\end{tabular}}} & \multicolumn{2}{|c}{\textbf{\begin{tabular}[c]{@{}c@{}}Remote\\ 
Objects\end{tabular}}} & \multicolumn{5}{|c}{\textbf{Metrics}} \\
\cmidrule{2-10}
 & \textbf{ G} & \textbf{P} & \textbf{ G} & \textbf{P} & \textbf{TL} & \textbf{NE(↓)} & \textbf{SR(↑)} & \textbf{OSR(↑)} & \textbf{SPL(↑)} \\
\midrule
 text &  &  &  &  & 19.1 & 5.97 & 46 & 60 & 37.0 \\
\midrule
text & $\checkmark$ &  &  &  & 17.1 & 4.61 & 66 & 76 & 53.2 \\
text &  & $\checkmark$ &  &  & 16.8 & 4.25 & 56 & 70 & 47.9 \\
\midrule
text & $\checkmark$ &  & $\checkmark$ &  & 18.4 & 6.51 & 60 & 72 & 48.7 \\
text &  & $\checkmark$ &  & $\checkmark$ & 18.3 & 5.37 & 56 & 70 & 47.9 \\
\midrule
visual & $\checkmark$ &  & $\checkmark$ &  & 14.9 & 4.34 & 72 & 74 & 58.8 \\
visual &  & $\checkmark$ &  & $\checkmark$ & 15.9 & 4.92 & 62 & 70 & 50.1 \\
\bottomrule
\end{tabular}}
\end{table}

\paragraph{Efficiency of compass-like visual observation.} 
As illustrated in Table~\ref{tab:ablation_compass}, among the compass-style representations, the 1024$\times$1024 resolution performs better. While the sequential input of eight directional views achieves the best overall performance, this observation format consumes over 1700 visual tokens per step, leading to high latency and cost. In contrast, our compass-style image representation requires only about 640 visual tokens at 1024$\times$1024 resolution, and can achieve navigation accuracy close to the sequential setting. Therefore, considering the efficiency, we adopt it as the default visual representation in SpatialNav.

\paragraph{Perception range of spatial map.} Table~\ref{tab:ablation_radius} demonstrates that the perception radius has a clear trade-off between the spatial information involved and the decision efficiency. A small radius of 3.84m, which is very close to the local perception range without using a spatial map, provides little extra information and therefore leads to limited performance improvement. Expanding the radius to 7.68m yields the best balance, achieving the strongest SR, SPL and nDTW. In contrast, an overly large radius of 11.52m introduces more distant, instruction-irrelevant rooms and navigable structures, which can dilute the agent attention. The SR slightly decreases despite a higher OSR. Therefore, we select 7.68m as the perception radius as it provides adequate spatial context without overwhelming redundancy. 


\section{Conclusions}

In this work, we propose to pre-explore the environment to construct a spatial scene graph about global spatial information. We then introduce SpatialNav, a zero-shot agent that effectively uses spatial scene graphs for navigation. Extensive experiments demonstrate that our method generalizes across environments and agents, enabling zero-shot performance comparable to learning-based methods.

\section*{Limitations}

While SpatialNav achieves robust zero-shot navigation performance, it still has several limitations. Firstly, how the spatial scene graphs can be integrated into supervised learning-based agents remains unexplored. Since most existing learning-based VLN agents are trained without access to top-down map-structured spatial representations, such inputs may not be compatible with these methods. We leave it as future work to collect additional VLN training data with spatial-map annotations and redesign the training paradigm to expose agents to this form of spatial information. 

\begin{table}[t]
\centering
\small
\captionsetup{skip=2pt}
\setlength{\tabcolsep}{2pt}
\setlength{\abovecaptionskip}{2pt}
\setlength{\belowcaptionskip}{3pt}
\caption{\textbf{Ablation on the visual format of panoramic observations.} 
``seq'' denotes that sequential images, whereas ``cps'' means a single compass-like image. }
\label{tab:ablation_compass}
\resizebox{\linewidth}{!}{
\begin{tabular}{c|c|c|cccc}
\toprule
\textbf{Fmt} & \textbf{\#Img} & \textbf{Img Size} & \textbf{OSR(↑)} & \textbf{SR(↑)} &  \textbf{SPL(↑)} &
\textbf{nDTW(↑)} \\
\midrule
cps & 1 & 1536$\times$1536 & 68.8   & 58.1  & 45.8  & 55.1 \\
cps & 1 & 1024$\times$1024 & 68.9   & \underline{60.3}  & \underline{50.1} & \underline{59.7} \\
cps & 1 & 512$\times$512   & \underline{69.7}  & 59.9  & 46.1  & 55.3 \\
\midrule
seq & 8 & 256$\times$256 & \textbf{70.0} &  \textbf{62.5}  & \textbf{54.6} & \textbf{63.3}  \\
\bottomrule
\end{tabular}}
\end{table}
\begin{table}[t]
\centering
\small
\captionsetup{skip=2pt}
\setlength{\tabcolsep}{3pt}
\setlength{\belowcaptionskip}{3pt}
\caption{\textbf{Ablation on the perception radius of agent-centric spatial map.} }
\label{tab:ablation_radius}
\begin{tabular}{c|cccccc}
\toprule
\textbf{Radius} & \textbf{OSR(↑)} & \textbf{SR(↑)} & \textbf{SPL(↑)} & \textbf{nDTW(↑)} \\
\midrule
11.52m  & \textbf{71.5}  & 56.9  & 47.0  & 58.43 \\
7.68m   & 68.9  & \textbf{60.3}  & \textbf{50.1}  & \textbf{59.7} \\
3.84m   & 62.2  & 50.9  & 42.9  & 55.33 \\
\bottomrule
\end{tabular}
\end{table}

Secondly, we admit that constructing a full spatial scene graph introduces additional computation overhead. Although modern SLAM systems can efficiently generate dense or sparse point clouds, we assume that these point clouds are available and do not evaluate the robustness and possible failure cases of the reconstruction process itself. Given that the reconstruction of 3D point cloud through pre-exploration can be computationally and operationally expensive, we consider fully evaluating and optimizing this pipeline as an important direction for future work. 
Furthermore, room-level segmentation remains challenging in open or ambiguously bounded spaces. Existing automated methods may produce inaccurate segmentation results, requiring extra manual correction. However, we find that this limitation is consistent with the situation in real-world robotic systems, where users are allowed to intervene after building the house map to refine the room layout.

\bibliography{papers}
\bibliographystyle{icml2026}

\end{document}